\documentclass{article}
\usepackage{makeidx,graphicx,times}
\begin{document}


\def\ap{\textrm{'}}
\def\Biops{{\sc Biops} }
\def\Biopsn{{\sc Biops}}
\def\Oops{{\sc Oops} }
\def\Oopsn{{\sc Oops}}
\def\Aixi{{\sc Aixi} }
\def\Aixin{{\sc Aixi}}
\def\tl{{\sc Aixi}{\em (t,l)} }
\def\tln{{\sc Aixi}{\em (t,l)}}
\def\hs{{\sc Hsearch} }
\def\hsn{{\sc Hsearch}}
\def\GM{G\"{o}del Machine }
\def\gm{G\"{o}del machine }
\def\GMn{G\"{o}del Machine}
\def\gmn{G\"{o}del machine}
\newtheorem{method}{Method}[section]
\newtheorem{principle}{Principle}
\newtheorem{procedure}{Procedure}[section]
\def\odt{{\textstyle{1\over 2}}}

\title{Simple Algorithmic Principles of Discovery, Subjective Beauty, 
Selective Attention, \\ Curiosity \& Creativity\thanks{Joint Invited 
Lecture for {\em Algorithmic Learning Theory} (ALT 2007) and 
{\em Discovery Science} (DS 2007), Sendai, Japan (preprint). Variant to 
appear in: 
V. Corruble, M. Takeda, and E. Suzuki (Eds.): DS 2007, 
LNAI 4755, pp. 26-38, Springer-Verlag Berlin Heidelberg 2007.
Abstract: M. Hutter, R.A. Servedio, and E. Takimoto (Eds.): 
ALT 2007, LNAI 4754, pp. 24-25, Springer-Verlag Berlin Heidelberg 2007; 
see also http://www.springerlink.com/content/y8j3453l0757m637/?p=42fb108af50a4cbf8ec06c12309884f6\&pi=2  and http://www.springer.com/west/home/generic/search/results?SGWID=4-40109-22-173760307-0}}

\date{}
\author{J\"{u}rgen Schmidhuber \\
TU Munich, Boltzmannstr. 3,  85748 Garching bei M\"{u}nchen, Germany \& \\
IDSIA, Galleria 2, 6928 Manno (Lugano), Switzerland \\
{\tt juergen@idsia.ch - http://www.idsia.ch/\~{ }juergen}}

\maketitle

\begin{abstract}

I postulate that human or other intelligent agents function or
should function as follows.  They store all sensory observations as 
they come---the data is `holy.'  At any time, given some agent's current
coding capabilities, part of the data is compressible by a short
and hopefully fast program / description / explanation / world
model. In the agent's subjective eyes, such data is more regular
and more {\em beautiful} than other data. It is well-known that
knowledge of regularity and repeatability may improve the agent's
ability to plan actions leading to external rewards.  In absence
of such rewards, however, {\em known} beauty is boring. Then {\em
interestingness} becomes the {\em first derivative} of subjective
beauty: as the learning agent improves its compression algorithm,
formerly apparently random data parts become subjectively more
regular and beautiful. Such progress in data compression is measured
and maximized by the {\em curiosity} drive: create action sequences
that extend the observation history and yield previously unknown /
unpredictable but quickly learnable algorithmic regularity. I
discuss how all of the above can be naturally implemented on
computers, through an extension of passive unsupervised learning
to the case of active data selection: we reward a general reinforcement
learner (with access to the adaptive compressor) for actions that
improve the subjective compressibility of the growing data.  An
unusually large compression breakthrough deserves the name
{\em discovery}.  The {\em creativity} of artists, dancers, musicians,
pure mathematicians can be viewed as a by-product of this principle.  
Several qualitative examples support this hypothesis.

\end{abstract}

\section{Introduction}
\label{intro}

A human lifetime lasts about $3 \times 10^{9}$ seconds.  The human
brain has roughly $10^{10}$ neurons, each with $10^4$ synapses
on average.  Assuming each synapse can store not more than 3 bits,
there is still enough capacity to store the lifelong sensory input
stream with a rate of roughly $10^5$ bits/s, comparable to the
demands of a movie with reasonable resolution. The storage capacity
of affordable technical systems will soon exceed this value.

Hence, it is not unrealistic to consider a mortal agent that interacts
with an environment and has the means to store the entire history
of sensory inputs, which partly depends on its actions. This data
anchors all it will ever know about itself and its role in the
world. In this sense, the data is `holy.'

What should the agent do with the data? How should it learn from it?  Which
actions should it execute to influence future data?

Some of the sensory inputs reflect external rewards. At any given
time, the agent's goal is to maximize the remaining reward or
reinforcement to be received before it dies. 
In realistic settings external rewards are rare though.
In absence of such rewards through teachers etc., what 
should be the agent's motivation?
Answer: It should spend some time on {\em unsupervised
learning}, figuring out how the world works, hoping this
knowledge will later be useful to gain external rewards.

Traditional unsupervised learning is about finding regularities,
by clustering the data, or encoding it through a
factorial code \cite{Barlow:89,Schmidhuber:92ncfactorial}
with statistically independent components, or
predicting parts of it from other parts.  All of this may be viewed
as special cases of data compression. For example, where there are
clusters, a data point can be efficiently encoded by its cluster
center plus relatively few bits for the deviation from the center.
Where there is data redundancy, a non-redundant factorial code 
\cite{Schmidhuber:92ncfactorial} will
be more compact than the raw data.  Where there is predictability,
compression can be achieved by assigning short codes to events
that are predictable with high probability \cite{Huffman:52}.  
Generally speaking we may say
that a major goal of traditional unsupervised learning is to improve
the compression of the observed data, by discovering a program that
computes and thus explains the history (and hopefully does so quickly) but 
is clearly shorter than the shortest previously known program of this kind.

According to our complexity-based theory of
beauty \cite{Schmidhuber:97art,Schmidhuber:98locoface,Schmidhuber:06cs},
the agent's currently achieved compression performance corresponds to 
subjectively perceived beauty: 
among several sub-patterns classified as `comparable' by a given 
observer, the subjectively most beautiful is the one with the simplest
(shortest) description, given the observer's particular method
for encoding and memorizing it.
For example, mathematicians find beauty in a simple proof with a
short description in the formal language they are using.
Others like geometrically simple, aesthetically pleasing, low-complexity
drawings of various objects \cite{Schmidhuber:97art,Schmidhuber:98locoface}.

Traditional unsupervised learning is not enough though---it 
just analyzes and encodes the data but does not choose it. We have
to extend it along the dimension of active action selection, since
our unsupervised learner must also choose the actions that influence
the observed data, just like a scientist chooses his experiments,
a baby its toys, an artist his colors, a dancer his moves,
or any attentive system its next sensory input.

Which data should the agent select by executing appropriate actions?
Which are the {\em interesting} sensory inputs that deserve
to be targets of its curiosity? I postulate 
\cite{Schmidhuber:06cs}
that in the absence of external
rewards or punishment the answer is: Those that 
yield {\em progress} in data compression. What does this mean?
New data observed by the learning agent may
initially look rather random and incompressible 
and hard to explain.
A good learner, however, will {\em improve} its 
compression algorithm over time, 
using some application-dependent learning algorithm,
making parts of the data history
subjectively more compressible, 
more explainable, more regular and more `beautiful.' 
A beautiful thing is interesting 
only as long as it is new, that is, 
as long as the algorithmic regularity 
that makes it simple has not yet been fully assimilated by 
the adaptive observer who is still learning to compress
the data better.
So the agent's goal should be: create action sequences
that extend the observation history and yield previously unknown /
unpredictable but quickly learnable algorithmic 
regularity or compressibility.  
To rephrase this principle in an informal way: 
maximize the {\em first derivative} of subjective
beauty. 

An unusually large compression breakthrough deserves the name
{\em discovery}.  How can we motivate a
reinforcement learning agent to make discoveries?  Clearly,
we cannot simply reward it for executing actions that 
just yield a compressible but boring history.  
For example,
a vision-based agent that always stays in the dark will experience
an extremely compressible and uninteresting history of unchanging 
sensory inputs.  Neither can we reward it for executing 
actions that yield highly informative but uncompressible data.
For example, our agent sitting in front of a 
screen full of white noise will experience highly unpredictable
and fundamentally uncompressible and uninteresting data 
conveying a lot of information in the traditional
sense of Boltzmann and Shannon \cite{Shannon:48}.
Instead, the agent should receive reward for creating / observing data that 
allows for {\em improvements} of the data's subjective compressibility.

The appendix will describe formal details of how to implement
this principle on computers.
The next section will provide 
examples of subjective beauty tailored to human observers,
and illustrate the learning process 
leading from less to more subjective beauty.
Then I will argue that the {\em creativity} of artists, dancers, musicians,
pure mathematicians as well as unsupervised {\em attention} in general
is just a by-product of our principle,  using
qualitative examples to support this hypothesis.

\section{Visual Examples of Subjective Beauty and 
its `First Derivative' Interestingness}

\begin{figure}[hbt]
\begin{minipage}[t]{1.00\textwidth}
\centerline{\includegraphics[height=6.2cm]{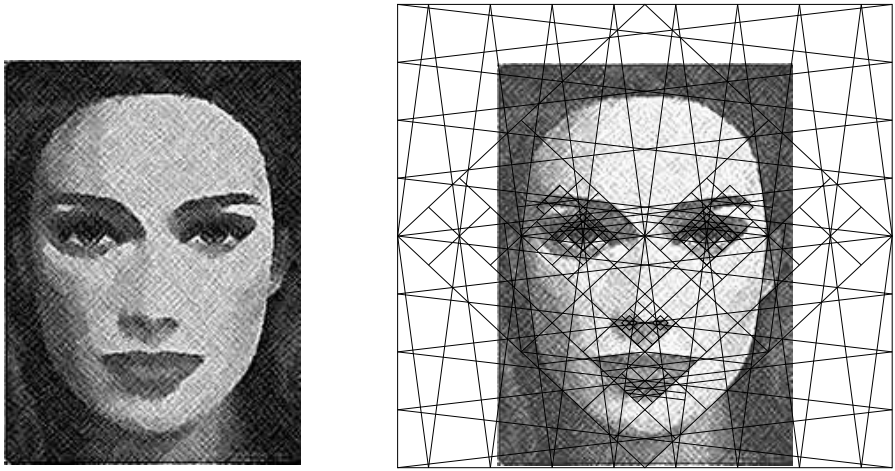}}
\end{minipage}
\caption{\em
{\bf Left:}
Drawing of a female face
based on a previously published construction plan \cite{Schmidhuber:98locoface} (1998).
Some human observers report they feel this face is `beautiful.'
Although the drawing has lots of noisy details (texture etc) without an 
obvious short description, positions and shapes of the basic facial
features are compactly encodable through a very simple geometrical scheme.
Hence the image contains a highly compressible algorithmic regularity or pattern 
describable by few bits of information.
An observer can perceive it through a sequence 
of attentive eye movements or saccades, and consciously or subconsciously discover 
the compressibility of the incoming data stream.
{\bf Right:} 
Explanation of how the essential
facial features were constructed  \cite{Schmidhuber:98locoface}.
First the sides of a square were partitioned into $2^4$ equal intervals.
Certain interval boundaries were connected to obtain
three rotated, superimposed grids based on lines with slopes
$\pm 1$ or $\pm 1/2^3$ or $\pm 2^3/1$.
Higher-resolution details of the grids were obtained 
by iteratively selecting
two previously generated, neighbouring, parallel lines and inserting
a new one equidistant to both.
Finally the grids were vertically compressed by a factor of $1-2^{-4}$.
The resulting lines and their intersections define
essential boundaries and shapes of eyebrows, eyes, lid shades,
mouth, nose, and facial frame in a simple way that 
is obvious from the construction plan.
Although this plan is simple in hindsight,
it was hard to find: hundreds of my previous attempts at discovering
such precise matches between simple geometries and pretty faces failed.
}
\label{locoface}
\end{figure}

Figure \ref{locoface} depicts the drawing of a female face considered
{\em `beautiful'} by some human observers. It also shows that the 
essential features of this face follow a very simple 
geometrical pattern \cite{Schmidhuber:98locoface} to
be specified by very few bits of information. That is,
the data stream generated by observing the image (say, through
a sequence of eye saccades) is more compressible than it would
be in the absence of such regularities. 
Although few people are able to immediately see 
how the drawing was made without studying its grid-based
explanation (right-hand side of Figure \ref{locoface}), most 
do notice that the facial features somehow fit together and
exhibit some sort of regularity. According to our postulate,
the observer's reward is generated by the conscious or 
subconscious discovery of this compressibility.
The face remains interesting
until its observation does not reveal any additional
previously unknown regularities. Then it becomes
boring even in the eyes of those who 
think it is beautiful---beauty and interestingness are two different things.

\begin{figure}[hbt]
\begin{minipage}[t]{0.49\textwidth}
\centerline{\includegraphics[height=6.25cm]{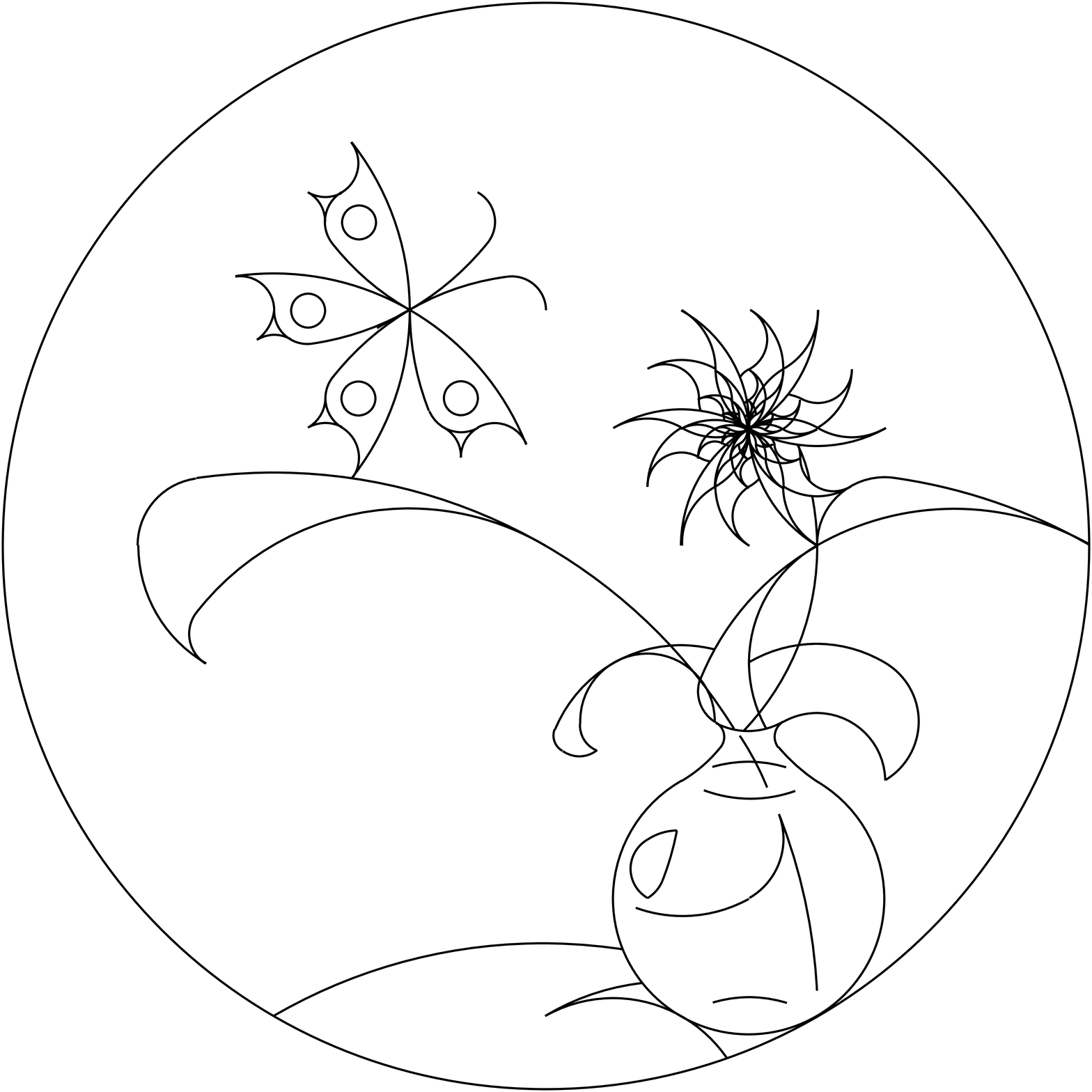}}
\end{minipage}
\hfill
\begin{minipage}[t]{0.49\textwidth}
\centerline{\includegraphics[height=6.25cm]{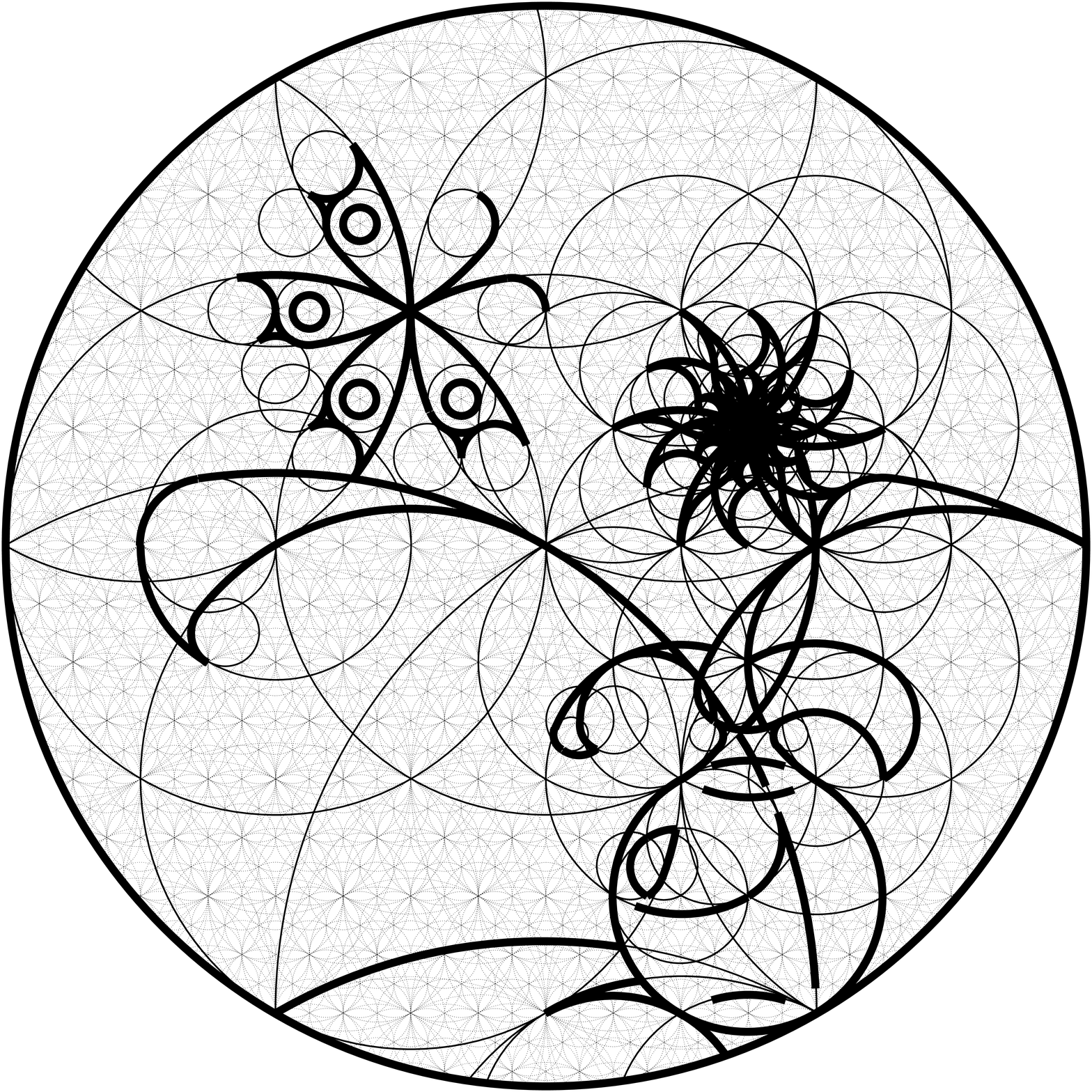}}
\end{minipage}
\caption{\em
{\bf Left:}
Image of a butterfly and a vase with a flower,
reprinted from {\em Leonardo}
\cite{Schmidhuber:97art,Schmidhuber:06cs}.
{\bf Right:} Explanation of how the image was constructed
through a very simple algorithm exploiting fractal circles
\cite{Schmidhuber:97art}.  The frame is a circle; its leftmost
point is the center of another circle of the same size.
Wherever two circles of equal size touch or intersect
are centers of two more circles with equal and half size, respectively.
Each line of the drawing is a segment of some circle, its
endpoints are where circles touch or intersect.
There are few big circles and many small ones. In general,
the smaller a circle, the more bits are needed to specify it.
The drawing to the left is simple (compressible) as it is based on few, rather large circles.
Many human observers report that they derive a certain amount of pleasure from
discovering this simplicity. The observer's learning process causes
a reduction of the subjective complexity of the data,
yielding a temporarily high derivative of subjective beauty.
(Again I needed a long time to discover
a satisfactory way of using fractal circles to
create a reasonable drawing.)
}
\label{butterfly}
\end{figure}

Figure \ref{butterfly}
provides another example: a butterfly and a vase with a flower.  
The image to the left can be specified by very few bits of
information; it can be constructed through a very simple procedure
or algorithm based on fractal circle patterns  
\cite{Schmidhuber:97art}.
People who understand this algorithm tend to appreciate
the drawing more than those who do not. 
They realize how simple it is. 
This is not an immediate,
all-or-nothing, binary process though. 
Since the typical human visual system has a lot of
experience with circles, 
most people quickly notice that the curves somehow 
fit together in a regular way. But few are
able to immediately state the precise 
geometric principles underlying the drawing. 
This pattern, however, is learnable from the 
right-hand side of Figure \ref{butterfly}.
The conscious or subconscious discovery process 
leading from a longer to a shorter
description of the data, or from less to more compression,
or from less to more subjectively perceived beauty, yields
reward depending on the first derivative of subjective beauty.

\newpage
\section{Compressibility-Based Rewards of Art and Music} 
\label{art}
The examples above indicate that 
works of art and music may have important purposes
beyond their social aspects \cite{sciencemusic:04}  
despite of those who classify art as superfluous \cite{Pinker:97}.
Good observer-dependent 
art deepens the observer's insights about this world or
possible worlds, unveiling previously unknown regularities
in compressible data,
connecting previously disconnected patterns 
in an initially surprising way that 
makes the combination of these patterns 
subjectively more compressible,
and eventually becomes 
known and less interesting. I postulate that the active
creation and attentive perception of all kinds of artwork 
are just by-products of my curiosity principle
yielding reward for compressor improvements.

Let us elaborate on this idea in more detail, following the discussion 
in \cite{Schmidhuber:06cs}.
Artificial or human  
observers must perceive art sequentially, and
typically also actively, e.g., through a sequence of attention-shifting
eye saccades or camera movements scanning a sculpture, or internal
shifts of attention that filter and emphasize sounds made by a
pianist, while surpressing background noise.
Undoubtedly many derive pleasure
and rewards from perceiving works of art, such as certain paintings,
or songs.  But different subjective observers 
with different sensory apparati and compressor improvement
algorithms will prefer different input sequences.
Hence any objective theory of what is good art must 
take the subjective observer as a parameter, to 
answer questions such as:
Which action sequences should he select to maximize
his pleasure?  According to our principle he
should select one
that maximizes the quickly learnable compressibility
that is new, relative to his current knowledge and his (usually
limited) way of incorporating or learning new data.

For example, which song should some human observer
select next?  Not the one
he just heard ten times in a row. It became too predictable in the
process.  But also not the new weird one with the completely
unfamiliar rhythm and tonality. It seems too irregular and contain
too much arbitrariness and subjective noise.  He should try 
a song that is unfamiliar enough to contain somewhat 
unexpected harmonies or melodies or beats etc., but
familiar enough to allow for quickly recognizing the presence of a
new learnable regularity or compressibility in the sound stream. 
Sure, this song will get boring over time, but not yet.

The observer dependence is illustrated by the fact that
Sch\"{o}nberg's twelve tone music is less popular than certain 
pop music tunes, presumably because its algorithmic structure is less
obvious to many human observers as it is based on more
complicated harmonies. For example, frequency ratios of successive
notes in twelve tone music often cannot be expressed as fractions 
of very small integers. 
Those with a prior education about 
the basic concepts and
objectives and constraints of twelve tone music, however, 
tend to appreciate Sch\"{o}nberg more than those without such an
education.

All of this perfectly fits our principle: 
The current compressor of a given subjective observer tries to compress 
his history of acoustic and other inputs where possible. The action selector
tries to find history-influencing actions that improve the compressor's performance on
the history so far.  The interesting musical and other subsequences are 
those with previously unknown yet learnable types of regularities, because they lead to
compressor improvements.  The boring patterns are those that seem arbitrary
or random, or whose structure seems too hard to understand.

Similar statements
not only hold for other dynamic art including
film and dance (taking into account the compressibility of controller actions), 
but also for painting and sculpture, which cause dynamic pattern sequences due to
attention-shifting actions \cite{SchmidhuberHuber:91} of the observer.

Just as observers get intrinsic rewards from
sequentially focusing attention on
artwork that exhibits new, previously unknown regularities,
the `creative' artists get reward for making it.  
For example, I found it extremely rewarding
to discover (after hundreds of frustrating failed attempts) the simple 
geometric regularities that permitted the construction of the
drawings in Figures \ref{locoface} and \ref{butterfly}.
The distinction between artists and observers is not clear though. 
Artists can be observers and vice versa. Both artists and observers execute action
sequences. The intrinsic motivations of both are fully compatible with
our simple principle.  
Some artists, however, crave {\em external} reward 
from other observers, in form of praise, money, or both, 
in addition to the {\em internal} reward that comes from creating a new work of art.
Our principle, however, conceptually separates these two types of reward.

From our perspective, scientists are very much like artists.
They actively select experiments in search for simple
laws compressing the observation history.
For example, different apples tend to fall off 
their trees in similar ways.
The discovery of a law underlying the acceleration
of all falling apples helps to greatly 
compress the recorded data.

The framework in the appendix is sufficiently formal
to allow for implementation of our principle on computers. 
The resulting artificial observers will vary in terms of
the computational power of their history compressors and learning algorithms.
This will influence what is good art / science to them, and what they find
interesting.  

\appendix

\section{Appendix}

This appendix is a compactified, compressibility-oriented
variant of parts of \cite{Schmidhuber:06cs}.

The world can be explained to a degree by compressing it.
The compressed version of the data can be viewed as its explanation.
Discoveries correspond to large data compression improvements 
(found by the given, application-dependent compressor improvement algorithm).
How to build an adaptive agent that not only tries to
achieve externally given rewards but also 
to discover, in an unsupervised and experiment-based fashion, 
explainable and compressible data?
(The explanations gained through explorative behavior may 
eventually help to solve teacher-given tasks.)

Let us formally consider a learning agent whose single life 
consists of discrete cycles or time steps $t=1, 2, \ldots, T$.
Its complete lifetime $T$ may or may not be known in advance.
In what follows, the value of any time-varying variable $Q$
at time $t$ ($1 \leq  t \leq T$) will be denoted by $Q(t)$,
the ordered sequence of values $Q(1),\ldots,Q(t)$ by $Q(\leq \! t)$,
and the (possibly empty) sequence $Q(1),\ldots,Q(t-1)$ by $Q(< t)$.
At any given $t$  the agent receives a real-valued input $x(t)$ from
the environment and executes a real-valued 
action $y(t)$ which may affect future inputs. At times $t<T$ its goal
is to maximize future success or {\em utility}
\begin{equation}
\label{u}
u(t) =
E_{\mu} \left [ \sum_{\tau=t+1}^T  
r(\tau)~~ \Bigg| ~~ h(\leq \! t) \right ],
\end{equation}
where $r(t)$ is an additional real-valued reward input at time $t$,
$h(t)$ the ordered triple $[x(t), y(t), r(t)]$
(hence $h(\leq \! t)$ is the known history up to $t$),
and $E_{\mu}(\cdot \mid \cdot)$ denotes the conditional expectation operator
with respect to some possibly unknown distribution $\mu$ from a set $\cal M$
of possible distributions. Here $\cal M$ reflects
whatever is known about the possibly probabilistic reactions
of the environment.  For example, $\cal M$ may contain all computable
distributions \cite{Solomonoff:64,Solomonoff:78,LiVitanyi:97,Hutter:04book+}.
There is just one life, no need for predefined repeatable trials, 
no restriction to Markovian 
interfaces between sensors and environment,
and the utility function implicitly takes into account the 
expected remaining lifespan $E_{\mu}(T \mid  h(\leq \! t))$
and thus the possibility to extend it through appropriate actions
\cite{Schmidhuber:05icann,Schmidhuber:05gmai,Schmidhuber:05gmconscious}.

Recent work has led to the first learning machines
that are universal and optimal in various very general senses 
\cite{Hutter:04book+,Schmidhuber:05icann,Schmidhuber:05gmai,Schmidhuber:03newai,Schmidhuber:06newmillenniumai,Schmidhuber:06ai75}.
Such machines can in principle find out by themselves whether
curiosity and
world model construction are useful or useless in a given
environment, and learn to behave accordingly.  
The present appendix, however, will assume {\em a priori} that
compression / explanation of the history is good and should be done;
here we shall not worry about the possibility
that `curiosity may kill the cat.'
Towards this end, in the spirit of our previous work
\cite{Schmidhuber:91singaporecur,Schmidhuber:91cur,Storck:95,Schmidhuber:97interesting,Schmidhuber:02predictable},
we split the reward signal $r(t)$ into
two scalar real-valued components: $r(t)=g(r_{ext}(t),r_{int}(t))$,
where $g$ maps pairs of real values to real values,  e.g., $g(a,b)=a+b$. 
Here $r_{ext}(t)$ denotes traditional {\em external} reward provided
by the environment, such as negative reward in 
response to bumping against a wall, or positive
reward in response to reaching some teacher-given goal state. 
But I am especially
interested in $r_{int}(t)$, the internal or intrinsic 
or {\em curiosity} reward, which is provided whenever
the data compressor / internal world model of the agent improves in
some sense. Our initial focus will be on
the case $r_{ext}(t)=0$ for all valid $t$.
The basic principle is essentially the one we published before in various variants
\cite{Schmidhuber:91cur,Schmidhuber:91singaporecur,Storck:95,Schmidhuber:97interesting,Schmidhuber:02predictable,Schmidhuber:04cur,Schmidhuber:06cs}:
\begin{principle}
\label{principle}
Generate curiosity reward for the controller in response to improvements of the history compressor.
\end{principle}
So we conceptually separate the goal (explaining / compressing the history)
from the means of achieving the goal. Once the goal is
formally specified in terms of an algorithm for computing curiosity rewards, 
let the controller's reinforcement learning (RL) mechanism figure 
out how to translate such rewards
into action sequences that allow the given compressor improvement algorithm
to find and exploit previously unknown types of compressibility.

\subsection{Predictors vs Compressors}
\label{predictor}

Most of our previous work on artificial curiosity
was prediction-oriented, e. g.,
\cite{Schmidhuber:91cur,Schmidhuber:91singaporecur,Storck:95,Schmidhuber:97interesting,Schmidhuber:02predictable,Schmidhuber:04cur,Schmidhuber:06cs}.
Prediction and compression are closely related though.
A predictor that correctly predicts 
many $x(\tau)$, given history $h(< \tau)$, for $1 \leq \tau \leq  t$,  
can be used to encode $h(\leq \! t)$ compactly: 
Given the predictor, only the wrongly predicted $x(\tau)$ plus
information about the
corresponding time steps $\tau$ are necessary
to reconstruct history $h(\leq \! t)$, e.g.,  \cite{Schmidhuber:92ncchunker}.
Similarly, a predictor that learns a probability distribution of
the possible next events, given previous events,  can be used to
efficiently encode observations with high (respectively low) predicted probability 
by few (respectively many) bits \cite{Huffman:52,SchmidhuberHeil:96}, thus achieving
a compressed history representation.
Generally speaking, we may view the predictor as the essential part of a 
program $p$ that re-computes $h(\leq \! t)$. 
If this program is short in comparison to the rad data $h(\leq \! t)$, then 
$h(\leq \! t)$ is regular or non-random 
\cite{Solomonoff:64,Kolmogorov:65,LiVitanyi:97,Schmidhuber:02ijfcs},
presumably reflecting essential environmental laws. Then
$p$ may also be highly useful for predicting future, yet unseen
 $x(\tau)$ for $\tau>t$.  

\subsection{Compressor Performance Measures}
\label{performance}
At any time $t$ ($1 \leq  t < T$), 
given some compressor program $p$ able to compress
history $h(\leq \! t)$, let $C(p,h(\leq \! t))$ denote 
$p$'s compression performance on $h(\leq \! t)$. 
An appropriate performance measure would be
\begin{equation}
C_l(p,h(\leq \! t))=l(p),
\end{equation}
where $l(p)$ denotes the
length of $p$, measured in number of bits: the shorter $p$,
the more algorithimic regularity and compressibility and 
predictability and lawfulness in the observations so far.
The ultimate limit for $C_l(p,h(\leq \! t))$ would be 
$K^*(h(\leq \! t))$, a variant of the Kolmogorov complexity
of $h(\leq \! t)$, namely, the length of the shortest program 
(for the given hardware) that computes an output
starting with $h(\leq \! t)$ 
\cite{Solomonoff:64,Kolmogorov:65,LiVitanyi:97,Schmidhuber:02ijfcs}.

$C_l(p,h(\leq \! t))$ does not take into account the time
$\tau(p,h(\leq \! t))$ spent by $p$ on computing $h(\leq \! t)$.
An alternative performance measure inspired by concepts
of optimal universal search \cite{Levin:73,Schmidhuber:04oops} is
\begin{equation}
C_{l \tau}(p,h(\leq \! t))= l(p) + \log~\tau(p,h(\leq \! t)).
\end{equation}
Here compression by one bit is worth as much as runtime
reduction by a factor of $\frac{1}{2}$.

\subsection{Compressor Improvement Measures}
\label{improvement}
The previous Section \ref{performance} only discussed measures
of compressor performance, but not of performance {\em improvement,}
which is the essential issue in our curiosity-oriented context.
To repeat the point made above:
 {\em The important thing are the improvements 
of the compressor, not its compression performance per se.}
Our curiosity reward in response to the 
compressor's progress 
(due to some application-dependent compressor improvement algorithm)
between times $t$ and $t+1$ should be
\begin{equation}
r_{int}(t+1)= f[C(p(t+1),h(\leq \! t+1)),C(p(t),h(\leq \! t+1))],
\end{equation}
where $f$ maps pairs of
real values to real values.  Various alternative progress measures
are possible; most obvious is $f(a,b)=a-b$.

Note that both the old and the new compressor have to be tested on the
same data, namely, the complete history so far.

\subsection{Asynchronous Framework for Creating Curiosity Reward}
\label{async}

Let $p(t)$ denote the agent's current compressor program at time $t$,
$s(t)$ its current controller, and do: 

\noindent \\
{\bf Controller:}
At any time $t$ ($1 \leq  t < T$) do:
\begin{enumerate}
\item
Let $s(t)$ use (parts of) history $h(\leq  t)$
to select and execute $y(t+1)$. 
\item
Observe $x(t+1)$.
\item
Check if there is non-zero curiosity reward $r_{int}(t+1)$
provided by the separate, asynchronously running
compressor improvement algorithm (see below).
If not, set $r_{int}(t+1)=0$.
\item
Let the controller's reinforcement learning (RL) algorithm use $h(\leq \! t+1)$
including $r_{int}(t+1)$
(and possibly also the latest available 
compressed version of the observed data---see below)
to obtain a new controller $s(t+1)$, 
in line with objective (\ref{u}).
\end{enumerate}

\noindent
{\bf Compressor:}
Set $p_{new}$ equal to the initial data compressor.
Starting at time 1, repeat forever until interrupted by death $T$:
\begin{enumerate}
\item
Set $p_{old}=p_{new}$; 
get current time step $t$ and set
$h_{old}=h(\leq \! t)$. 
\item
Evaluate $p_{old}$ on $h_{old}$, to obtain $C(p_{old},h_{old})$
(Section \ref{performance}).
This may take many time steps.
\item
Let some (application-dependent)
compressor improvement algorithm 
(such as a learning algorithm for
an adaptive neural network predictor)
use $h_{old}$
to obtain a hopefully better compressor $p_{new}$ 
(such as a neural net with the same size but
improved prediction capability
and therefore improved compression performance). 
Although this may take many time steps, $p_{new}$ 
may not be optimal, due to limitations of
the learning algorithm, e.g., local maxima.
\item
Evaluate $p_{new}$ on $h_{old}$, to obtain $C(p_{new},h_{old})$.
This may take many time steps.
\item
\label{fasync}
Get current time step $\tau$ and  generate curiosity reward 
\begin{equation}
r_{int}(\tau)=f[C(p_{old},h_{old}),C(p_{new},h_{old})], 
\end{equation}
e.g., $f(a,b)=a-b$; see Section \ref{improvement}.
\end{enumerate}
Obviously this asynchronuous scheme
may cause long temporal delays 
between controller actions and corresponding 
curiosity rewards. This may impose a heavy
burden on the controller's RL algorithm whose task
is to assign credit to past actions
(to inform the controller about beginnings of compressor 
evaluation processes etc., 
we may augment its input by unique representations of such events).
Nevertheless, there are 
RL algorithms for this purpose which are
theoretically optimal in various senses,
to be discussed next.

\subsection{Optimal Curiosity \& Creativity \& Focus of Attention}
\label{optimalcur}

Our chosen compressor class typically will have
certain computational limitations.  In the absence of any external rewards,
we may define {\em optimal pure curiosity
behavior} relative to these limitations:
At time $t$ this behavior would select the action that maximizes
\begin{equation}
\label{optcur}
u(t) = E_{\mu} \left [ \sum_{\tau=t+1}^{T} 
r_{int}(\tau)~~ \Bigg| ~~ h(\leq \! t) \right ].
\end{equation}
Since the true, world-governing
probability distribution $\mu$ is unknown, 
the resulting task of the controller's RL algorithm 
may be a formidable one. 
As the system is revisiting previously uncompressible parts of the environment,
some of those will tend to  become more compressible, that is, the corresponding
curiosity rewards will
decrease over time. A good RL algorithm must somehow
detect and then {\em predict} this decrease, and act accordingly.  
Traditional RL algorithms \cite{Kaelbling:96}, however,
do not provide any theoretical guarantee of optimality for such
situations. 
(This is not to say though that sub-optimal
RL methods may not lead to success in certain applications; 
experimental studies might lead to interesting insights.)

Let us first make the natural assumption that the compressor 
is not super-complex such as Kolmogorov's, that is, its output 
and $r_{int}(t)$ are computable for all $t$. 
Is there a best possible RL algorithm that comes as
close as any other to maximizing
objective (\ref{optcur})? Indeed, there is. 
Its drawback, however, is that it is not
computable in finite time. Nevertheless,
it serves as a reference point for defining
what is achievable at best.

\subsection{Optimal But Incomputable Action Selector}
\label{aixi}

There is an optimal way of selecting actions
which makes use of Solomonoff's theoretically optimal
universal predictors and their Bayesian learning algorithms
\cite{Solomonoff:64,Solomonoff:78,LiVitanyi:97,Hutter:04book+,Hutter:07uspx}.
The latter only assume that the reactions of the environment are sampled from 
an unknown probability distribution $\mu$ contained in a set $\cal M$
of all enumerable distributions---compare text after equation (\ref{u}).
More precisely, given an observation sequence $q(\leq \! t)$,
we only assume there exists a computer program
that can compute the probability of the next possible $q(t+1)$, given 
$q(\leq \! t)$. In general we do not know this program, hence
we predict using a mixture distribution 
\begin{equation}
\label{xi}
\xi(q(t+1)\mid q(\leq \! t)) =\sum_i w_i\mu_i (q(t+1)\mid q(\leq \! t)),
\end{equation}
a weighted sum of {\em all} distributions $\mu_i \in \cal M$, $i=1, 2, \ldots$, 
where the sum of the constant weights satisfies $\sum_i w_i \leq 1$. 
This is indeed the best one can possibly do, in a very general sense
\cite{Solomonoff:78,Hutter:04book+}. The drawback
of the scheme is its incomputability, since $\cal M$ contains
infinitely many distributions.
We may increase the theoretical power of the scheme by
augmenting $\cal M$ by certain
non-enumerable but
limit-computable distributions \cite{Schmidhuber:02ijfcs},
or restrict it such that it becomes computable,
e.g., by assuming the world is computed
by some unknown but deterministic computer
program sampled from the Speed Prior \cite{Schmidhuber:02colt} which assigns
low probability to environments that are hard to compute by any method.

Once we have such an optimal predictor, we can extend it
by formally including the effects of executed actions to define an 
optimal action selector maximizing future expected reward.
At any time $t$, Hutter's theoretically optimal
(yet uncomputable) RL algorithm \Aixi \cite{Hutter:04book+} 
uses an extended version of Solomonoff's prediction scheme  
to select those action sequences that promise maximal
future reward up to some horizon $T$,  given the current data $h(\leq \! t)$.
That is, in cycle $t+1$, \Aixi
selects as its next action the first action of an action sequence
maximizing $\xi$-predicted reward up to the given horizon, appropriately
generalizing eq. (\ref{xi}).
\Aixi uses observations optimally \cite{Hutter:04book+}: 
the Bayes-optimal policy $p^\xi$ based on
the mixture $\xi$ is self-optimizing in the sense that its average
utility value converges asymptotically for all $\mu \in \cal M$ to the
optimal value achieved by the Bayes-optimal policy $p^\mu$
which knows $\mu$ in advance. The necessary and sufficient condition is
that $\cal M$ admits self-optimizing policies.
The policy $p^\xi$ is also Pareto-optimal
in the sense that there is no other policy yielding higher or equal
value in {\em all} environments $\nu \in \cal M$ and a strictly higher
value in at least one \cite{Hutter:04book+}.

\subsection{Computable Selector of Provably Optimal Actions, Given Current System}
\label{gm}

\Aixi above needs unlimited computation time. Its computable variant
\tl \cite{Hutter:04book+} has asymptotically optimal runtime but may suffer
from a huge constant slowdown.
To take the consumed computation time into account in a general,
optimal way, we may use the recent \gmn s 
\cite{Schmidhuber:05icann,Schmidhuber:05gmai,Schmidhuber:05gmconscious}
instead. They
represent the first class of mathematically rigorous, fully
self-referential, self-improving, general, optimally efficient problem solvers.
They are also applicable to the problem embodied by 
objective (\ref{optcur}).

The initial software  $\cal S$ of such a \gm contains
an initial problem solver, e.g., 
some typically sub-optimal  method
\cite{Kaelbling:96}.
It also contains an asymptotically optimal  initial proof
searcher based on an online variant of Levin's
{\em Universal Search} \cite{Levin:73},
which is used to run and test {\em proof techniques}. Proof techniques
are programs written in a universal language implemented
on the \gm within $\cal S$. They are in principle  able to compute proofs
concerning the system's own future performance, based on an axiomatic 
system $\cal A$ encoded in $\cal S$.
$\cal A$ describes the formal {\em utility} function, in our case eq. (\ref{optcur}),
the hardware properties, axioms of arithmetic and probability
theory and data manipulation etc, and  $\cal S$ itself, which is possible
without introducing circularity
\cite{Schmidhuber:05gmai}.

Inspired by Kurt G\"{o}del's celebrated self-referential formulas (1931),
the \gm rewrites any part of its own code (including the proof searcher) through
a self-generated executable program as soon
as its {\em Universal Search} variant has found a proof that the rewrite is {\em useful} 
according to objective (\ref{optcur}).
According to the Global Optimality Theorem
\cite{Schmidhuber:05icann,Schmidhuber:05gmai,Schmidhuber:05gmconscious},
such a self-rewrite is globally optimal---no local maxima possible!---since
the self-referential code first had to prove that it is not useful to continue the 
search for alternative self-rewrites.

If there is no provably useful optimal
way of rewriting $\cal S$ at all, then humans
will not find one either. But if there is one,
then $\cal S$ itself can find and exploit it.  Unlike the previous {\em
non}-self-referential methods based on hardwired proof searchers \cite{Hutter:04book+},
\gmn s not only boast an optimal {\em order} of complexity but can optimally
reduce (through self-changes) any slowdowns hidden by the $O()$-notation, provided the utility
of such speed-ups is provable.

\subsection{Consequences of Optimal Action Selecton}
\label{consequences}
Now let us apply any optimal RL algorithm to curiosity rewards as defined above.
The expected consequences are: at time $t$ the controller will do the best to
select an action $y(t)$
that starts an action sequence expected to create observations
yielding maximal expected compression {\em progress} up to the expected
death $T$, taking into accunt the limitations of both the
compressor and the compressor improvement algorithm.  
In particular, ignoring issues of computation time, 
it will focus in the best possible way on things
that are currently still uncompressible but will soon become
compressible through additional learning.
It will get bored by things that already are compressible.
It will also get bored by things
that are currently uncompressible but will apparently
remain so, given the experience so far,
or where the costs of making
them compressible exceed those of making other
things compressible, etc.


\bibliography{bib}
\bibliographystyle{plain}
\end{document}